\documentclass[letter-size paper,10pt]{article} 
\usepackage[american]{babel}
\usepackage{amsmath}
\usepackage{amssymb}
\usepackage{algorithmicx,algorithm}
\usepackage[noend]{algpseudocode}

\usepackage[noend]{algpseudocode}

\usepackage[utf8]{inputenc}
\usepackage{authblk}
\usepackage{upgreek}
\usepackage{spconf}
\usepackage{indentfirst}
\usepackage{subfigure}
\usepackage{graphicx}
\usepackage{geometry}
\usepackage{amsthm}
\usepackage{txfonts}
\usepackage{float}

\usepackage{mathtools} 
\usepackage{booktabs} 
\usepackage{tikz} 
\usepackage{cite}
\usepackage{natbib} 
\title{Thompson Sampling on Asymmetric $\alpha$-Stable Bandits}

\begin{document}
    
	\name{Zhendong Shi, Ercan E. Kuruoglu, Xiaoli Wei}
	               \address{Tsinghua-Berkeley Shenzhen Institute 
	               \\
		                        shizd20@mails.tsinghua.edu.cn, kuruoglu@sz.tsinghua.edu.cn, xiaoli\_wei@sz.tsinghua.edu.cn}
	\date {}

\maketitle

\begin{abstract}
  In algorithm optimization in reinforcement learning, how to deal with the exploration-exploitation dilemma is particularly important. Multi-armed bandit problem can optimize the proposed solutions by changing the reward distribution to realize the dynamic balance between exploration and exploitation. Thompson Sampling is a common method for solving multi-armed bandit problem and has been used to explore data that conform to various laws. In this paper, we consider the Thompson Sampling approach for multi-armed bandit problem, in which rewards conform to unknown asymmetric $\alpha$-stable distributions and explore their applications in modelling financial and wireless data.


\end{abstract}

\section{Introduction}\label{sec:intro}
        Sequential decision-making plays a key role in many fields, such as quantitative finance and robotics. In order to make real-time decisions under unknown environments, decision makers must carefully design algorithms to balance the trade-off between exploration and exploitation. Many decision algorithms have been designed and widely used, such as financial decision-making \cite{Mahapatro} and personalized news recommendation.
		
		The multi-armed bandits (MAB) have an important potential in solving the dilemma of exploration and exploitation in sequential decision-making problem in which a fixed limited set of resources must be allocated between competing (alternative) choices in a way that maximizes their expected gain. With different reward distributions, it can deal with different data. Through the research on this problem for many years, for some common reward distribution functions from Bernoulli distribution and Gaussian distribution to sub-exponential family, fast processing algorithms like termed UCB-Rad \cite{Multi-armed} (specifically for MAB with sub-exponential rewards) have been developed.

		However, when we design decision-making algorithms for complex systems, the reward distribution function (such as Bernoulli distribution and Gaussian distribution) is inconsistent with the probability distribution that each arm actually obeys. According to the research on these complex system data, one can observe that interactions often lead to heavy-tailed and power law distribution \cite{Frequency}, such as modeling stock prices, social networks, and online behavior on websites. These extensions from Gaussian distribution to more complex and practical reward distribution allow for the opportunity to develop significantly more efficient algorithms than were possible in the original setting as long as we capture reward distribution in many real world applications.
		
		The $\alpha$-stable distributions is a family of distribution with power law heavy tails, which can better fit the actual reward distribution and be applied to the exploration of complex systems. This family refers to a class of distributions parameterized by the exponent $\alpha$.
		
		Existing machine learning algorithms are difficult to deal with the problem of multi armed bandits with complicated reward distributions. This is because the probability density of such reward distributions, such as $\alpha$-stable distribution, can not be obtained analytically. Under some characteristics of complex real data, such as under impulsive or unsymmetric characteristics the standard algorithms probably choose the incorrect arms.
		
		In view of the above problems, \cite{Thompson} have studied the symmetric $\alpha$-Thompson Sampling method to provide a mode that fits the non-Gaussian data with impulsive characteristics better and make better decisions. Another alternative principle motivated by the Upper Confidence Bound (UCB) Algorithm has emerged as a practical competitor to make the best choice in the recommendation system \cite{Recommendation}. We will also compare these algorithms to verify the validity of our algorithm.

		2. Our work provides the finite-time polynomial regret bound on the BayesianRegret of Thompson Sampling achieved by asymmetric $\alpha$-TS in this setting.
		
		3. Through a series of experiments, the proficiency of our algorithm is proved by comparison with greedy algorithm, UCB algorithm and symmetric $\alpha$-TS algorithm in the case of asymmetric $\alpha$-stable reward.

\section{BACKGROUND MATERIAL}
The multi-armed bandit \cite{Finite} is a mathematical model that have been used widely in machine learning and optimization, and a number of algorithms have been proposed to optimally solve the bandit problem when the reward distributions are Gaussian-distribution or exponential-distribution \cite{Nathaniel} . However, these reward distributions could not cover all conditions when designing algorithms for complex systems. When we model stock prices or deal with behaviour in social networks, the interactive data often lead to heavy tail and negative skewness \cite{skewness}. \cite{Thompson} propose symmetric $\alpha$-Thompson Sampling method to fit heavy tailed data. In our cases, we use asymmetric $\alpha$-Thompson Sampling method to fit data with negative skewness distribution.

\subsection{Multi-armed bandit problem}
        Suppose that there are several slot machines placed in front of an agent. For each round, we can select a slot machine to pull and record the rewards given by the slot machine. Assuming that each slot machine is not exactly the same, after multiple rounds of operation, we can mine some statistical information of each slot machine, and then select the slot machine that seems to have the highest reward.

        In any case of multi-armed bandit problem with N arms, there is an agent who can access a set of N actions (or "arms"). Learning is carried out in rounds and indexed by t $\in$ [T]. The total number of rounds called time range T is known in advance. This problem is iterative, the agent picks arm $a_t$ $\in$ [N] and then observes reward $r_{a_t}$ (t) from that arm in each round of t $\in$ [T].
		
		For each arm n $\in$ [N], rewards come from a distribution $D_n$ with mean $\mu_n$ = $E_{D_n} [r]$.
		The largest expected reward is denoted by $\mu^{\star}$ = $\max_{n \in [N]}$ $\mu_n$ , and the corresponding arm(s) is denoted as the optimal arm(s) $n^*$.
		
		It is impractical to simply take the total reward as the standard to measure the quality of the strategy. Therefore, we define regret R(T) as an indicator of good or bad strategy, which refers to the difference between the most ideal total reward the agent can achieve and the total reward the agent actually gets.

        \begin{equation}
			\begin{aligned}
				R(T) =  \mu^{\star}T - \sum_{t=0}^{T} \mu_{a_t}
			\end{aligned}
			\label{f1}
		\end{equation}

\subsection{Thompson Sampling}
There are various exploration algorithms, including from the initial greedy algorithm that randomly selects the arm with probability $\epsilon$ and selects the best arm so far with probability $1-\epsilon$, to the UCB algorithm that increases the exploration probability of the unconvinced part with inefficient random exploration, and then to the Thompson sampling using Bayesian method instead of frequency. 
		
		Compared with UCB method, Thompson sampling makes greater use of prior knowledge by conjugate prior. In the MAB problem, the most basic Bernoulli distribution happens to have the beta distribution as the conjugate prior. With the conjugate priors, it is easy to do posterior update. To be exact, in MAB, each time the recommended item is selected, the parameters of beta distribution are updated according to whether the item is recommended. The beta distribution with new parameters is a more accurate estimation of the conversion rate of the selected items after combining this experience. 
		
		Suppose $f$ is the probability density function of beta distribution, both $\alpha$ and $\beta$ are positive parameters, also known as shape parameters, $x \in (0,1)$.
		\begin{equation}
			\begin{aligned}
				f(x,\alpha,\beta) = \frac{1}{B(\alpha,\beta)} x^{\alpha-1} (1-x)^{\beta-1}
			\end{aligned}
			\label{f2}
		\end{equation}
		
		After studying the algorithm when the reward distribution conforms to Bernoulli distribution, we consider extending it to the general case. We assume that for each arm n, the reward distribution is $D_n$, and the parameter $\theta_n$ comes from a set of $\Theta$, with a prior probability distribution p($\theta_n$), and Thompson sampling algorithm selects the arm based on the a posterior probability of the reward under the arm. For each round of t $\in$ [T], the agent draws the parameter $\hat{\theta}_n$(t) for each arm n $\in$ [N] from the posterior distribution of parameters. 
        \begin{equation}
			\begin{aligned}
				\hat{\theta}_n(t) \sim p(\theta_n|r_n(t-1)) \propto p(r_n(t-1) | \theta_n) p(\theta_n)
			\end{aligned}
			\label{f3}
		\end{equation}

       Given the drawing parameter $\hat{\theta}_n$(t) of each arm, the agent selects the arm $a_t$ with the largest average return on the posterior distribution, taking action $a_t$ for return $r_t$, and updates arm action $a_t$ a posterior distribution.
		\begin{equation}
			\begin{aligned}
				a_t = \mathop{\arg\max}\limits_{n \in [N]} \mu_n (\theta_n(t))
			\end{aligned}
			\label{f4}
		\end{equation}
		
		We will use the Bayes Regret for the measure for performance. Bayes Regret is the expected regret over the priors.Denoting the parameters over all arms as $\overline{\theta} = \{ \theta_1 , ..., \theta_n \}$, their corresponding product distribution as $ \overline{D} = \prod_i D_i$.
		\begin{equation}
			\begin{aligned}
				Bayesian Regret(T,\pi) = E_{\overline{\theta} \sim \overline{D}} [R(T)]
			\end{aligned}
			\label{f5}
		\end{equation}

\subsection{$\alpha$-stable distribution}
An important generalization of the Gaussian distribution which can model both impulsive and skewed data is the
		$\alpha$-stable distribution which can be described with its characteristic function:
		\begin{equation}
			\begin{aligned}
				\phi_\eta(u)  = \exp\{-\sigma^{\alpha} \left|u\right|^{\alpha} (1-i \beta sign(u) \times \tan \frac{\pi \alpha}{2}) + i u \mu \} 
			\end{aligned}
			\label{f6}
		\end{equation}
	    Where  $\alpha$ is the characteristic exponent defining the impulsiveness of the distribution, $\beta$ is the symmetry parameters, $\gamma$ is the scale parameter and $\mu$ is the location parameter. 
		
		We denote $\eta$ as $S_\alpha(\sigma, \beta, \mu)$. In our paper, we concentrate on the interval $\alpha$ $\in$ (1,2) because there is not even a finite first-order moment when $\alpha$ $\textless$ 1, mathematical expectations are not available.
		
        Most of the work on $\alpha$-stable distribution applications consider only the symmetric case. However, various work point out  asymmetric $\alpha$-stable laws as in \cite{herranz04},  \cite{kuruoglu03},  \cite{nguyen19}, \cite{xu20}.

\subsection{Symmetric $\alpha$-Thompson Sampling}
Considering the setting, for each arm n, the corresponding reward distribution is given by $D_n = $ $S_\alpha(\sigma, \beta, \mu_{n})$ where $\alpha \in$ (1,2), $\sigma \in \mathbb{R}^+$ are known in advance as say in \cite{Thompson} study, and $\mu_n$ is unknown. Note that for symmetric $\alpha$ stable distribution ($\beta = 0$),  $E[r_n]=\mu_n$, and hence we set a prior distribution over the variable $\mu_n$ which is the expected reward for arm n. We can see that since the only unknown parameter for the reward distributions is $\mu_n$ ,$D_n$ is parameterized by $\theta_n$ = $\mu_n$.
		
		As we can see, the only unknown parameter of reward distribution is $\mu_n$ , therefore, $D_n$ is parameterized by $\theta_n$= $\mu_n$. For each round of t $\in$ [T], agent draws parameters $\hat{\theta}_n$(t) = $\overline{\mu}_n$(t) for arm n $\in$ [N] from the posterior distribution of parameters which is given by the previous rewards $r_n$(t - 1) = $\{ r_n^{(1)}$, $r_n^{(2)}$, ...$\}$, gives the drawn parameters $\overline{\mu}_n$(t)) and obtains largest reward $r_t$ after taking action $a_t$ and then updates the posterior distribution for arm $a_t$.
		
		This algorithm is called Symmetric $\alpha$-Thompson Sampling \cite{Thompson}.

\subsection{Bootstrapped UCB}
In addition to the symmetric $\alpha$-Thompson Sampling algorithm, we also present the greedy algorithm \cite{Greedy} and UCB algorithm which we will use for comparison in our experiments.
		
		The greedy algorithms rigidly divide selection process into exploration stage and exploration stage. During exploration, all items are explored with the same probability without using any historical information. 
		
		In order to realize exploration and development faster and better, we need to make full use of historical information. Each time we want to choose the best arm, if an arm has been recommended N times (N feedback has been obtained), the probability of choosing correct arm can be calculated. In reality, the number of recommended arms cannot be infinite. Therefore, there is always a difference between the estimated probability of correct arm and the probability of correct real arm. Based on the above two information, we can define a new strategy: every time we recommend, we are always optimistic that the return that the arm can get is the sum of valuation and difference. This is the famous upper confidence bound (UCB) algorithm. The optimistic aspect of this algorithm comes from the fact that despite being uncertain about the estimated means of the distributions of the $N$ arms, at each time step $t$ it always selects the arm $n$ that maximizes the sum $r_n+a(n,t)$, where $r_n$ is the estimated mean of machine $n$ and $a(n,t)$ is the UCB of machine $n$ at time step $t$.
		
		Definition: $a(n,t)$ is the upper confidence bound of machine $n$ at time step $t$, $p_{n,t}$ is the number of times that machine $n$ has been pulled at time step $t$.
		$a(n,t) = \sqrt{\frac{2log(t)}{p_{n,t}}}$
        
\section{Asymmetric $\alpha$-Thompson Sampling}
As we mentioned in the introduction, our improvement on the symmetric $\alpha$-Thompson Sampling algorithm is that we consider the influence of the negative skewness in the data on the algorithm results.

\subsection{Location parameter for $\alpha$-stable process}
		
		Since we use $\mu$ as an unknown parameter in the algorithm, different from the symmetric case, we need to solve the problem that the $\mu$  cannot be directly substituted by the mean value in the asymmetric case by constructing a new $\mu$.
		
		For the asymmetric condition ($\beta \neq 0$), we can also connect the location parameter and reward on the basis of properties of stable distribution. We establish this connection through following Theorem \cite{Roberto}.
		
		Theorem 1: Let $X_n$ become independent identically distributed stable random variables with parameter $\alpha, \beta, \sigma, \mu$. $X_n \sim S_{\alpha}(\sigma,\beta,\mu)$ and $a_n \in R$, n $\in$ [N], then 
		\begin{equation}
			\begin{aligned}
				Z &= \sum_{n=1}^{N}a_n X_n \sim
 \\ &= S_{\alpha}((\sum_{n=1}^{N}|a_n|^{\alpha})^{1/{\alpha}}\sigma,\frac{\sum_{n=1}^{N}a_n^{<\alpha>}}{\sum_{n=1}^{N}|a_n|^{\alpha}}\beta,\sum_{n=1}^{N}a_n\mu)
			\end{aligned}
			\label{f7}
		\end{equation}
		where $a_n^{<\alpha>} = sign(a_n)|a_n|^{\alpha}$.
		
		We can derive Lemma 1 which shows the transformation of $\alpha$-stable random variables from asymmetric one to symmetric one using the Theorem 1.
		
		Lemma 1: Let $X_n \sim S_{\alpha}(\sigma,\beta,\mu)$, n $\in$ [N] and $X_n^D \sim X_{3k} + X_{3k-1} - 2^{1/\alpha}X_{3k-2}$, then $X_n^D \sim S_{\alpha}(4^{1/\alpha}\sigma,0,[2-2^{1/\alpha}]\mu)$.
		
		$X^D$ is non-skewed transformation, $\mu^D$ is its mean-value, thus,
		\begin{equation}
			\begin{aligned}
				\hat{\mu} = \hat{\mu^D}(2-2^{1/\alpha})^{-1}
			\end{aligned}
			\label{f8}
		\end{equation}
		
		Estimate $\hat{\mu}$ of parameter $\mu$ is unbiased and consistent.
		
		\subsection{Algorithm for asymmetric $\alpha$-Thompson Sampling}
		
		By parameter estimation \cite{estimation} of the initial data, we get the parameters $\alpha, \beta, \sigma$ and prior distribution for $\mu$.
		
		Suppose a set of observation data $\{x_1, x_2, ..., x_N\}$ follows a stable distribution and suppose the characteristic function is $\phi$.
		\begin{equation}
			\begin{aligned}
				\phi(u)= E[e^{ius}] \approx \frac{1}{N} \sum_{j=1}^N e^{iu{s_j}}
			\end{aligned}
			\label{f9}
		\end{equation}

		\begin{equation}
			\begin{aligned}
				\phi(u)= e^{-|\sigma u|^{\alpha}} (\cos\eta+i*\sin\eta)
			\end{aligned}
			\label{f10}
		\end{equation}
        where $\eta$ = $\sigma u -|\sigma u|^{\alpha}\beta sgn(u) \tan\frac{\alpha \pi}{2}$
        
		\begin{equation}
			\begin{aligned}
				\hat{\alpha}= \frac{\log\frac{\log|\hat{\phi}(u_1)|}{\log|\hat{\phi}(u_2)|}}{\log|\frac{u_1}{u_2}|}
			\end{aligned}
			\label{f11}
		\end{equation}
		
		\begin{equation}
			\begin{aligned}
				\log \hat{\sigma}= \frac{\log|u_1|\log(-\log|\hat{\phi}(u_2)|)-\log|u_2|\log(-\log|\hat{\phi}(u_1)|)}{\log|\frac{u_1}{u_2}|}
			\end{aligned}
			\label{f12}
		\end{equation}
		
		Assume that $\psi(u):= \mu u - |\sigma u|^{\alpha}\beta sign(u)tan \frac{\alpha \pi}{2}$
		\begin{equation}
			\begin{aligned}
				\hat{\beta}= \frac{u_4 \psi(u_3)-u_3 \psi(u_4)}{\hat{\sigma}^{\hat{\alpha}}\tan\frac{\pi \hat{\alpha}}{2}(u_4 u_3^{\hat{\alpha}}-u_3 u_4^{\hat{\alpha}})}
			\end{aligned}
			\label{f13}
		\end{equation}
		
		According to the assumption that the initial data conforms to the $\alpha$ stable distribution and the obtained $\alpha, \beta, \sigma$ through parameter estimation, we can deduce the probability distribution of the last parameter $\mu$ value.
		
		Then, through Bayesian inference, which is aimed at updating prior parameter knowledge, in the form of density function $p(\theta)$, for symmetric stable variables, $\theta = \mu $, while for asymmetric stable variables the relationship between mean reward and location parameter, we could choose $\theta = \mu^D$ to simplify the function $p(\theta)$.
		
		Using observable data r and parametric function $f(r|\theta)$, the result is given by $p(\theta|r) \propto f(r|\theta) p(\theta)$. 
		
		Due to the complicated probability density functionl of the stable distribution, especially for the asymmetric one, we need to introduce some extra random variables y, then the updating process can be decomposed via the integration formula $p(\theta|r) \propto \int f(r,y|\theta) p(\theta) dy$.
		
		Theorem 2: Let $f$ be the bivariate probability density function of $\hat{Z}$ and $\hat{Y}$, conditional on $\alpha$ and $\beta$.
		\begin{equation}
			\begin{aligned}
				f(z,y|\alpha,\beta,\sigma, \mu, x) = \frac{\alpha}{|\alpha-1|}  \exp\{-|\frac{z}{t_{\alpha,\beta}(y)}|^{\alpha/(\alpha-1)}\} \\ \times |\frac{z}{t_{\alpha,\beta}(y)}|^{\alpha/(\alpha-1)} \frac{1}{|z|}
			\end{aligned}
			\label{f14}
		\end{equation}
		where $t_{\alpha,\beta}(y)$ = $(\frac{\sin[\pi \alpha y+\eta_{\alpha,\beta}]}{\cos \pi y})(\frac{\cos \pi y }{\cos [\pi (\alpha-1) y + \eta_{\alpha,\beta}]})^{\alpha/(\alpha-1)}$ and $\eta_{\alpha,\beta} = \beta \min (\alpha,2-\alpha) \pi /2$ and $l_{\alpha,\beta} = -\frac{\eta_{\alpha,\beta}}{\pi \alpha}$. Then $f$ is a proper bivariate probability density for distribution of (Z,Y), and maginal distribution of Z is $S_{\alpha}(\beta,0,1)$.
		
		According to Theorem 2, we derive the posterior density of asymmetric alpha-stable distribution conditional on $\alpha$, $\beta$ and $\sigma$ by following the equation.
		
		We begin with the density of $f(y|\alpha,\beta,\sigma,\mu,x)$:
		\begin{equation}
			\begin{aligned}
				f(y|\alpha,\beta,\sigma,\mu,x) \propto \exp\{-|\frac{z}{t_{\alpha,\beta}(y)}|^{\alpha/(\alpha-1)}+1\} \\
|\frac{z}{t_{\alpha,\beta}(y)}|^{\alpha/(\alpha-1)}
			\end{aligned}
			\label{f15}
		\end{equation}
		where z = $\frac{x-\mu}{\sigma}$.
		
		By the auxiliary random variable y, we have that
		\begin{equation}
			\begin{aligned}
				p (\mu | \alpha, \beta, \sigma, x, y) \propto \exp \{ -\sum_{i=1}^N  |\frac{z_i}{t_{\alpha,\beta}(y_i)}|^{\alpha/(\alpha-1)}  \} \\
				\times \prod_{i=1}^N 
				\frac{z_i}{t_{\alpha,\beta}(y_i)}|^{\alpha/(\alpha-1)} \frac{1}{|x_i-\mu|} p(\mu)
			\end{aligned}
			\label{f16}
		\end{equation}
		where $z_i = \frac{x_i- \mu}{\sigma}$ and $p (\mu)$ is the prior distribution for $\mu$.

		\begin{algorithm}[H]
        \caption{Asymmetric $\alpha$-Thompson Sampling}
        \hspace*{0.02in} {\bf Input:}
        Arms n $\in$ [N], priors $\alpha, \beta, \sigma$ for each arm, auxiliary variable y
		\begin{algorithmic}[1]
        \State set $\theta$ = $\mu^D$ = $\mu$($2$-$2^{\frac{1}{\alpha}}$) as estimated parameter
        \For{each iteration t $\in$ [T]} 
        \For{each arms n $\in$ [N]}
            \State Draw $\hat{\theta_n(t)} \sim \pi(\theta_n|r_n(t-1)) \newline$
            $\propto \int f(r_n(t-1),y|\theta_n) \pi(\theta_n) dy$ \newline
\State $f(r_n(t-1),y|\theta_n) \propto  $\newline $\exp\{-|\frac{z}{t_{\alpha,\beta}(y)}|^{\alpha/(\alpha-1)}+1\}|\frac{z}{t_{\alpha,\beta}(y)}|^{\alpha/(\alpha-1)}$
\State $\pi(\theta_n|\alpha,\beta,\sigma,r,y) \propto \newline \exp \{-\sum_{i=1}^N |\frac{z_i}{t_{\alpha,\beta}(y_i)}|^{\alpha/(\alpha-1)}\} 
				\newline \times \prod_{i=1}^N 
				\frac{z_i}{t_{\alpha,\beta}(y_i)}|^{\alpha/(\alpha-1)} \frac{1}{|r_i-\theta_n|} \pi(\theta_n)$
\EndFor
\EndFor

\end{algorithmic}
\end{algorithm}

		\subsection{Regret Analysis}
		
		Suppose the loss function is $g$, which represents the loss function corresponding to the round of data. We represents the parameters corresponding to the prediction model when the wheel data is used.
		
		Upper bound can not only used in UCB method, but also on the finite-time Bayesian Regret \cite{Upper} (BR) incurred by the asymmetric $\alpha$-Thompson Samplings algorithm. 
		
		Continuing with the theorem used, we assume a N armed bandit with T maximum trials. Each arm n follows an $\alpha$-stable reward $S_{\alpha}(\beta,\sigma,\mu_n)$, let $\mu^{\star} = \max_{n \in [N]} \mu_n$ denote the arm with maximum reward. The Bayesian regret is derived by the following theorem.
		
		Theorem 3: (Regret Bound). Let N $\textgreater$ 1, $\alpha \in (1,2)$, $\sigma \in \mathbb{R}^+$, $\mu_{n \in [N]} \in [-M,M]$. For a N-armed bandit with rewards for each arm n drawn from $S_{\alpha}(\beta,\sigma,\mu_n)$, for $\epsilon$ chosen a priori such that $\epsilon \rightarrow (\alpha-1)^{-}$,
		\begin{equation}
			\begin{aligned}
				BayesianRegret(T,\pi^{TS}) = O(N^{\frac{1}{1+\epsilon}}T^{\frac{1+\epsilon}{1+2\epsilon}})
			\end{aligned}
			\label{f17}
		\end{equation}
		
		Proof. Let $k_n c$ denote the number of times arm n has been pulled until time t while N-armed bandit with rewards for arm n are drawn from $S_{\alpha}(\beta,\sigma,\mu_n)$, represents the experience average reward of arm n before time t by $r_n(t)$, represents the arm pulled by t as $a_t$ at any time, and the best arm represents $a_t^{*} $. Upper confidence bound for arm n at any time t is
		\begin{equation}
			\begin{aligned}
				U_k(t) = clip_{[-M,M]} [r_k(t)+\sigma^{\frac{1}{\alpha}}(\frac{2C(1+\epsilon,\alpha)}{n_k(t)^{\epsilon}})^{\frac{1}{1+\epsilon}}]
			\end{aligned}
			\label{f18}
		\end{equation}
		for $0 \le \epsilon \le 1$, $M \ge 0$. Let E be the event when for all k $\in$ [K] arms, over all iterations t $\in$ [T], we have:
		\begin{equation}
			\begin{aligned}
				|r_k(t)-\mu_k| \le \sigma^{\frac{1}{\alpha}}(\frac{2C(1+\epsilon,\alpha)}{ n_k(t)^{\epsilon}})^{\frac{1}{1+\epsilon}}
			\end{aligned}
			\label{f19}
		\end{equation}
	
		Compared to the problem-independent regret bound of $O(\sqrt{N T \log(T)})$ for Thompson Sampling on the multi-armed Gaussian bandit problem, our bound has a super-linear dependence of order $T^{\frac{1+\epsilon}{1+2\epsilon}}$. 

\section{Data}
		
		Our test sets include asymmetric data generated using Chambers Mallows Stuck algorithm \cite{Chambers}, individual stock cases like AIV, from yfinance package in Python and wireless Network data provided by \cite{Game}.
		
		We choose synthetic asymmetric data to test the efficiency of asymmetric $\alpha$-TS algorithm and log return of closing stock price from 2016/07/01 to 2020/07/18, wireless Network data consisted in the latencies measured when visiting Internet home pages of 760 universities to show the application of asymmetric $\alpha$-TS algorithm.

		\begin{figure} [H]
			
			\flushleft
			\includegraphics[width=0.5\textwidth]{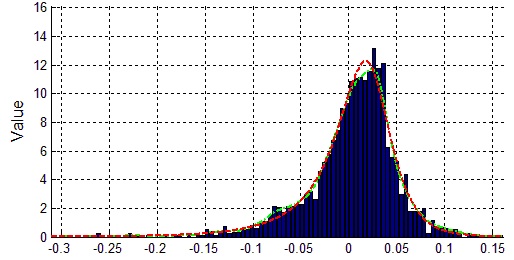}
			\label{fig2}
			\caption{stock price data, blue histogram represents the log return of stock price data provided by the yfinance, the red curve represents the fitted symmetric $\alpha$ stable distribution and the green one represents the fitted asymmetric $\alpha$ stable distribution, skeness is the main difference between symmetric and asymmetric data, and symmetric $\alpha$-stable distribution can not perfectly fit asymmetric data.}

		\end{figure}
		
\begin{table}[H]
    \centering
    \caption{Parameter estimation for asymmetric stable distribution}\label{tab:data}
    \begin{tabular}{|c|c|c|c|c|}
      \toprule 
      Tscode & Theta & Alpha & Beta & Sigma \\
      \midrule 
      Asymmetric data & 0.0000602 & 1.32 & -0.13 & 0.0006\\
      Financial data & 0.0000771 & 1.38 & -0.21 & 0.0011\\
      Wireless data & 0.0000171 & 1.72 & -0.32 & 0.0011\\
      \bottomrule 
    \end{tabular}
\end{table}
		
		This data shows the log return of the stock price, which conforms to asymmetric $\alpha$-stable distribution due to its negative skewness and large kurtosis.

\section{Experimental Study}
		In order to show the efficiency and stability of the asymmetric $\alpha$-TS algorithm in a specific data field, we will use the $\epsilon$-greedy algorithm, bootstrapped UCB algorithm, symmetry $\alpha$-TS algorithm and asymmetric algorithm in different data sets for comparative experiments. The number of sampling iterations is set as 2000.
		
		\subsection{Synthetic asymmetric data}
		
		This asymmetric data set is generated using Chambers Mallows Stuck algorithm \cite{Chambers}. We conducted multi-armed bandit experiments on above benchmarks, where are $\alpha$= 1.3 and K = 30 arms respectively. The average value of each arm is randomly selected , and the $\sigma$= 500 of each experiment. Each experiment was run for t = 2000 iterations, and the average regret shows below.
		 
		 \begin{figure} [H]
			
			\centering
			\includegraphics[width=0.5\textwidth]{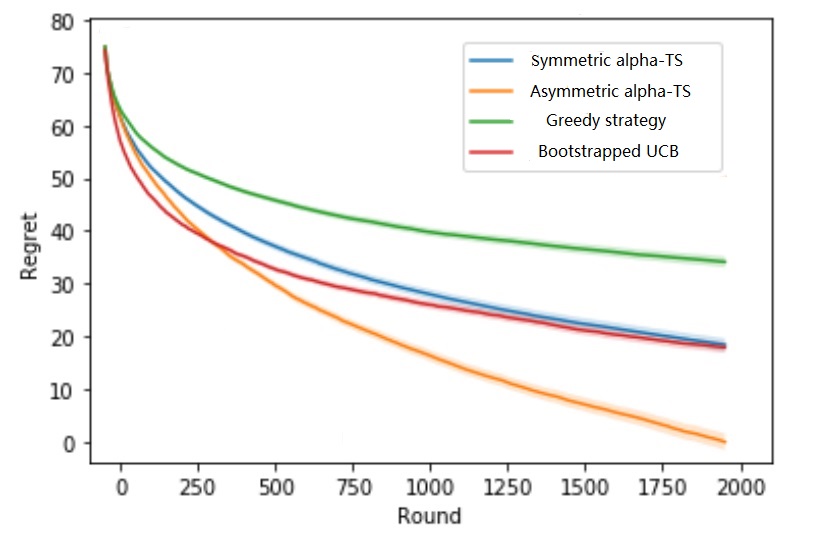}
			\label{fig3}
			\caption{regret for asymmetric data, the green line is greedy strategy, blue one is common alpha-TS method, red one is Bootstrapped UCB while orange line shows our asymmetric $\alpha$-TS method}
			
		\end{figure}
		The test results of the asymmetric data meet our expectations, and the symmetric algorithm performs worse than our method in time and space efficiency. The figures summarizes the types of bandit problems used to solve the machine-learning problems mentioned above. 
		\subsection{Stock selection}
		
		We consider a stochastic Stock portfolio selection problem in order to illustrate the versatility of Thompson sampling for bandit settings more complicated than just one original data each time.

		\begin{figure} [H]
			
			\centering
			\includegraphics[width=0.5\textwidth]{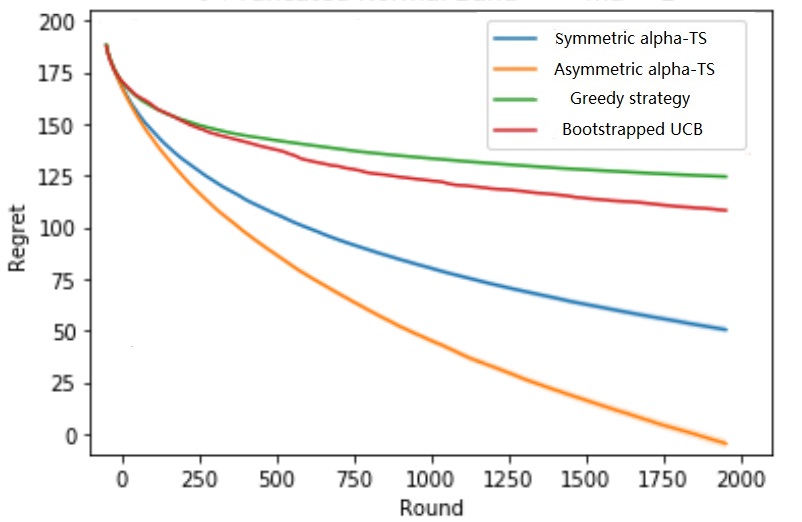}
			\label{fig4}
			\caption{Regret for stock price data, the green line is greedy strategy, blue one is common alpha-TS method, red one is Bootstrapped UCB while orange line shows our asymmetric $\alpha$-TS method}
			
		\end{figure}
		
		Every stock has a different, unknown mean duration, with the stock means taken to be equally spaced in [0, N]. At each round, one stock arrives to the selector, with a random duration that follows the exponential distribution with the corresponding mean.
		
		According to the test results and parameter estimation of test data, it can be inferred that the greater the skewness, the more obvious the gap between the performances of symmetric algorithm and the asymmetric algorithm. 
		\subsection{WIRELESS COMMUNICATIONS DATA}
		
		The data included measured delays in accessing the Internet home pages of 760 universities. Delay to 1361 × 760 matrix, where each column represents the delay measured when collecting data samples from one of the web pages.
		
		\begin{figure} [H]
			
			\centering
			\includegraphics[width=0.5\textwidth]{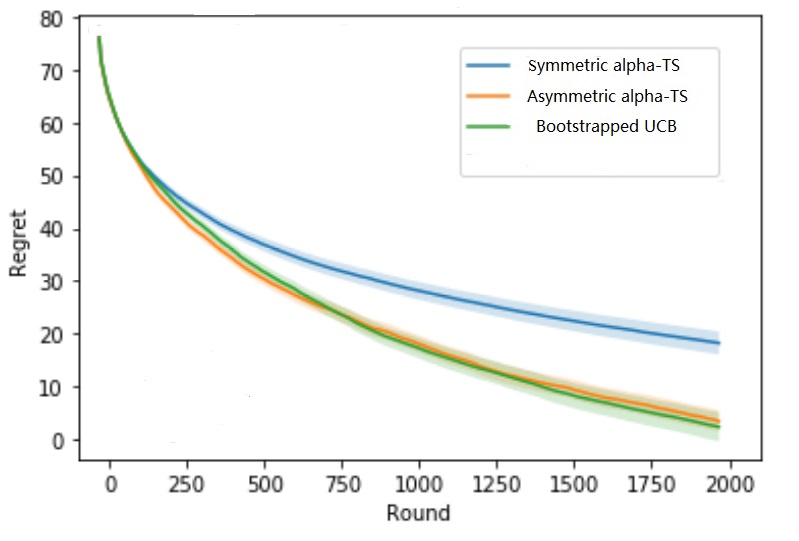}
			\label{fig5}
			\caption{regret for Wireless Network data, the green line is Bootstrapped UCB, blue one is common alpha-TS method while orange line shows our asymmetric $\alpha$-TS method}

		\end{figure}
		 The relative success of UCB algorithm lies in its use of historical information. UCB does not use randomness at all. In each case, the arm selected by UCB can be calculated from data. Therefore, in the face of highly correlated data such as social information, the effect of UCB algorithm is better than other algorithms, while in the face of more random generated data and financial data, its advantages are difficult to show.
		
		The results given in this section clearly show that the best choice to use the algorithm depends largely on the allocation of rewards. Conservative algorithms that spend more time measuring/exploring, such as greedy strategy, should be the preferred choice when one arm is significantly better than others, while our algorithm should be more inclined to use when different arms have similar rewards.
		
		\section{Conclusion}
	
		In view of the complexity of action/observation space in many problems, we design an asymmetric $\alpha$-Thompson sampling algorithm using Bayesian inference for stable distribution and verify the conjecture through synthetic asymmetric data and real stock price data. For the data consistent with their respective conjectures, both symmetric algorithm and asymmetric algorithm show their advantages over other algorithms. Asymmetric algorithm can also be used to process symmetric data because it has no restrictions on beta, but because it uses complicated Bayesian inference formula (In the symmetry $\alpha$-Thompson algorithm, the operation process can be greatly simplified through the characteristic of symmetry and alternative variables), the iteration speed can not be compared with symmetric $\alpha$-Thompson algorithm which can iterate from prior distribution to posterior distribution immediately under symmetric conjecture and auxiliary variables.
		
		Based on the regret bound for symmetric $\alpha$ Thompson sampling, we develop a regret bound for asymmetric $\alpha$ one in the parameter, action and observation spaces. Regret bound can be used to determine the rationality of the data and even apply it to other algorithms such as Robust $\alpha$ Thompson sampling, which assumes that for all arms, reward can only be feedback as truncated mean estimator for each arm, any arm is discarded for each iteration i if it is larger than the bound.
		
		Applying the algorithm to stock price returns, telecommunication and recommendation system, we prove that asymmetric stable distribution is a better data model, which can explain the existence of skewness and kurtosis. If necessary, the degree of deviation from normal can be evaluated from the following samples. In addition, our intermediate concentration results provide a starting point for other machine learning problems for financial or behavioral data that may be investigated under the stable setting of $\alpha$.

\bibliography{uai2022}

\end{document}